\documentclass[conference]{IEEEtran}

\IEEEoverridecommandlockouts

\usepackage{amsmath,amsfonts,bm}

\def\eqref#1{equation~\ref{#1}}

\def\1{\bm{1}}

\DeclareMathAlphabet{\mathsfit}{\encodingdefault}{\sfdefault}{m}{sl}
\SetMathAlphabet{\mathsfit}{bold}{\encodingdefault}{\sfdefault}{bx}{n}

\newcommand{\gray}[1]{\textcolor{gray}{#1}}

\newcommand{\conditionalcomment}[1]{\if\commenttext1 \else {#1} \fi}
\newcommand{\grayconditionalcomment}[1]{\if\commenttext1 \else \gray{{#1}} \fi}

\renewcommand{\eqref}[1]{Eq.~(\ref{#1})}

\makeatletter
\DeclareRobustCommand\onedot{\futurelet\@let@token\@onedot}
\def\@onedot{\ifx\@let@token.\else.\null\fi\xspace}

\makeatother

\usepackage{stfloats}
\usepackage{dsfont}
\usepackage{multirow}
\usepackage{url}
\usepackage{xspace}
\usepackage[accsupp]{axessibility}
\usepackage{physics}
\usepackage{graphicx}
\usepackage{lipsum} 
\usepackage{booktabs}
\usepackage{scalerel}
\usepackage{algorithm}
\usepackage{pifont}

\usepackage{colortbl}
\definecolor{blue(ncs)}{rgb}{0.0, 0.53, 0.74}
\definecolor{green(ncs)}{rgb}{0.0, 0.62, 0.42}
\definecolor{cadmiumorange}{rgb}{0.93, 0.53, 0.18}

\usepackage[dvipsnames]{xcolor}

\definecolor{grey}{rgb}{0.9, 0.9, 0.9}
\newcommand{\ccol}{\cellcolor{grey}}

\usepackage{tabularx}
\usepackage[export]{adjustbox}

\newcommand{\smallbreakparagraph}[1]{\smallbreak \noindent \textbf{#1}}

\usepackage[hidelinks]{hyperref}

\usepackage{cite}
\usepackage{amsmath,amssymb,amsfonts}
\usepackage{algorithmic}
\usepackage{graphicx}
\usepackage{textcomp}
\usepackage{xcolor}
\def\BibTeX{{\rm B\kern-.05em{\sc i\kern-.025em b}\kern-.08em
    T\kern-.1667em\lower.7ex\hbox{E}\kern-.125emX}}

\begin{document}

\title{Component Adaptive Clustering for Generalized Category Discovery}

\author{\IEEEauthorblockN{Mingfu Yan\textsuperscript{1,2*}, Jiancheng Huang\textsuperscript{1*}\thanks{* Both authors contributed equally to this research.}, Yifan Liu\textsuperscript{1}, Shifeng Chen\textsuperscript{1,3$\dagger$}\thanks{$\dagger$ Corresponding author. This work was supported by the Shenzhen Science and Technology Program (JSGG20220831105002004) and Shenzhen Key Laboratory of Computer Vision and Pattern Recognition.}}
\IEEEauthorblockA{\textsuperscript{1}Shenzhen Institutes of Advanced Technology, Chinese Academy of Sciences, Shenzhen, China \\
\textsuperscript{2}University of Chinese Academy of Sciences, Beijing, China\\
\textsuperscript{3}Shenzhen University of Advanced Technology, Shenzhen, China
}
}

\maketitle

\newcommand{\shortmethod}{AdaGCD}

\begin{abstract}
Generalized Category Discovery (GCD) tackles the challenging problem of categorizing unlabeled images into both known and novel classes within a partially labeled dataset, without prior knowledge of the number of unknown categories. Traditional methods often rely on rigid assumptions, such as predefining the number of classes, which limits their ability to handle the inherent variability and complexity of real-world data. To address these shortcomings, we propose \textit{\shortmethod}, a cluster-centric contrastive learning framework that incorporates Adaptive Slot Attention (AdaSlot) into the GCD framework. AdaSlot dynamically determines the optimal number of slots based on data complexity, removing the need for predefined slot counts. This adaptive mechanism facilitates the flexible clustering of unlabeled data into known and novel categories by dynamically allocating representational capacity. By integrating adaptive representation with dynamic slot allocation, our method captures both instance-specific and spatially clustered features, improving class discovery in open-world scenarios. Extensive experiments on public and fine-grained datasets validate the effectiveness of our framework, emphasizing the advantages of leveraging spatial local information for category discovery in unlabeled image datasets.
\end{abstract}


\section{Introduction}
\label{sec:introduction}

Given an image dataset where only a portion of the images are labeled, the model must predict the categories for the remaining unlabeled images. Notably, these predicted categories may include novel classes not present in the labeled dataset, a challenge referred to as Generalized Category Discovery (GCD) \cite{vaze2022gcd}. Deep learning \cite{lv2024gpt4motion,huang2024sbcr,li2024comae,huang2024magicfight,li2024sglp,huang2024dive,li2025fedkd,huang2024color,huang2024wavedm,xiong2025enhancing,huang2024diffusion,luo2023wizardmath,luo2024wizardarena} has revolutionized computer vision, with methods like contrastive learning enabling models to learn from large amounts of unlabeled data.
To address it, some researchers have advocated for directly training the feature extractor with labeled and unlabeled data within a contrastive learning framework~\cite{vaze2022gcd}. Some approaches have proposed generating high-quality pseudo-labels for the unlabeled data~\cite{cst, zhou2019collaborative, tieredimagenet, s3d}.

Although the aforementioned methods have shown promising results in the GCD task, they face limitations when applied to testing images, as they rely solely on global category features and ignore the spatial local information of the test image itself. In reality, objects within images are characterized by more than just global information. An object can be understood as a rich semantic assembly of multiple components, each contributing distinct features such as color, texture, and form. For example, a cat comprises elements like the soft fur-covered head, pointed ears with a velvety texture, the padded paws with tiny claws, and the long, sleek tail that might be striped or solid in color. Current methods, which encode images only from a global category feature perspective, neglect a substantial amount of spatial local information, creating a bottleneck in this task.

In addition to global category features, we posit spatial local features (such as texture and color) are essential. 
These spatial local features are retained in spatial local information with 2D spatial structures (e.g., the final layer local features extracted by DINO~\cite{dino}, which preserve an \( H \times W \) structure). While current GCD methods tend to discard all local features, we aim to leverage this spatial local information for GCD by introducing a contrastive learning framework that operates from image-scale perspectives.

\begin{figure}[t]
    \centering
    \includegraphics[width=1\linewidth]{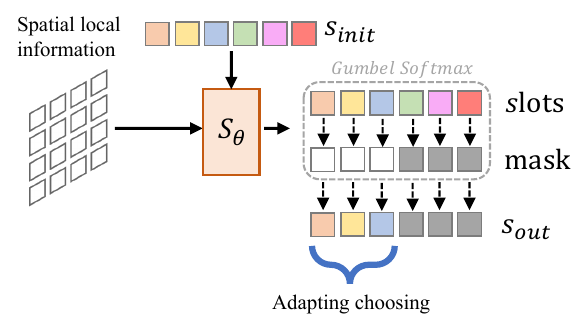}
    \caption{Illustration of the Adaptive Slot Attention~\cite{fan2024adaptive}.}
    \vspace{-0.5cm}
    \label{fig:adaslot}
\end{figure}

\begin{figure*}[t]
    \centering
    \includegraphics[width=\linewidth]{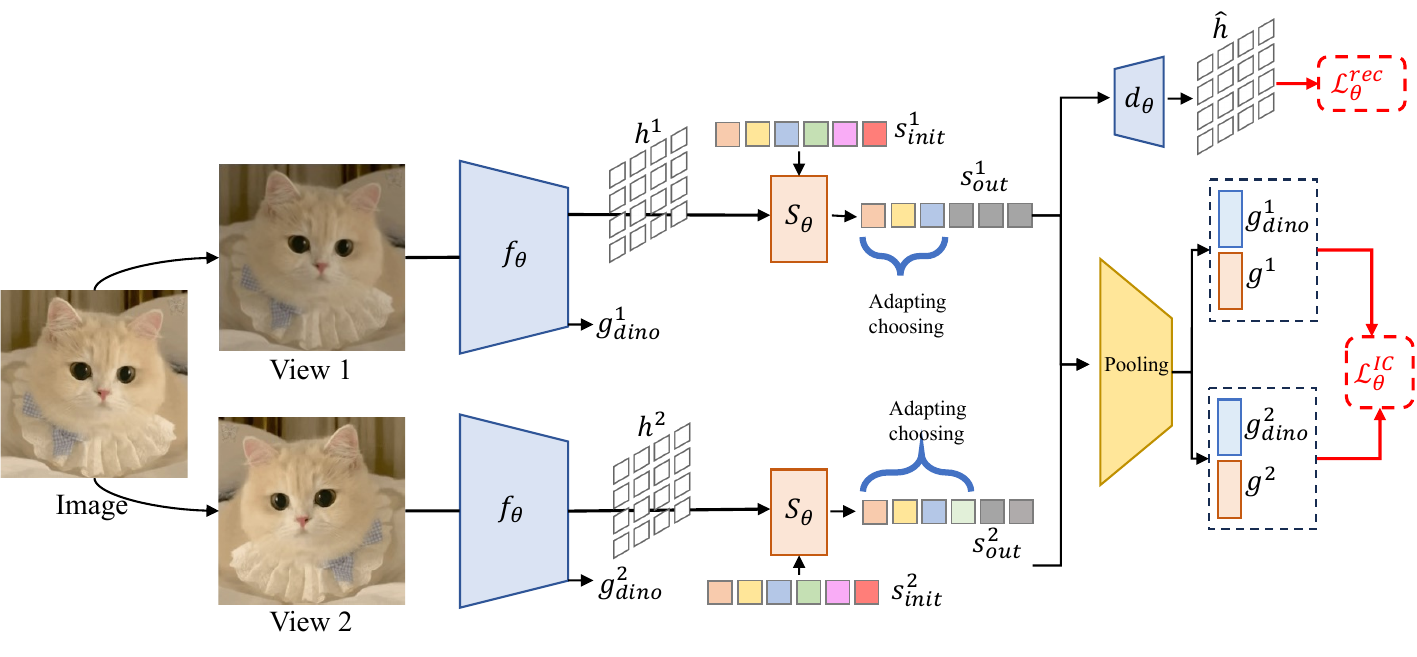}
    \caption{
This figure illustrates our proposed method, \shortmethod, a cluster-centric contrastive learning framework for Generalized Category Discovery (GCD). Each augmented view of the input image is processed through a shared encoder $f_{\theta}$, which generates spatial local feature maps $\{h^1, h^2\}$ and global representations $\{g^1_{\mathrm{dino}}, g^2_{\mathrm{dino}}\}$. The component clusterer module $\mathcal{S}_{\theta}$, initialized with $\{s^{1}_{\text{init}}, s^{2}_{\text{init}}\}$, captures clustered features, producing $\{s^1_{\mathrm{out}}, s^2_{\mathrm{out}}\}$. A masked slot decoder~\cite{fan2024adaptive} $d_{\theta}$ (indicated in blue) manages reconstruction tasks, while the pooled outputs—comprising both global representations and clustered features—are utilized for image-scale contrastive loss $\mathcal{L}^{\mathrm{IC}}_{\theta}$. This unified approach integrates spatial local and global representations, enabling robust category discovery.
    }
    \label{fig:framework}
    \vspace{-0.3cm}
\end{figure*}

We propose a cluster-centric contrastive learning framework designed to capture both global and spatial local features. To extract clustered features, the framework employs Adaptive Slot Attention (AdaSlot), as illustrated in Fig.~\ref{fig:adaslot}, to decompose spatial local feature maps into multiple semantic components, with each slot representing a distinct spatial local feature. The image-scale representation is then derived by pooling over these slot features, providing the input for the final image-scale contrastive learning stage.
The main contributions of this paper are as follows:  
1) We develop a cluster-centric contrastive learning framework, as shown in Fig. \ref{fig:framework}, that effectively leverages the spatial local information of images.  
2) We integrate a slot attention mechanism into the contrastive learning framework, enabling the model to explicitly learn slot-scale representations.  
3) We conduct extensive experiments on various image datasets, demonstrating the effectiveness of the proposed method.

\section{Related Work}
\label{sec:related_work}

\subsection{Generalized Category Discovery (GCD)}

The Generalized Category Discovery (GCD) task \cite{vaze2022gcd} extends the Novel Category Discovery (NCD) framework \cite{ncd} by addressing scenarios where unlabeled images belong to both known and unknown classes. In contrast, NCD assumes that all unlabeled images originate from unknown classes. This distinguishes GCD from Semi-supervised Learning (SSL) \cite{cst, zhou2019collaborative, tieredimagenet, s3d}, which presumes that all unlabeled data belong exclusively to known classes. GCD introduces a more realistic setting, where some unlabeled data may correspond to classes not present in the labeled set, similar to NCD. However, unlike NCD—which aggregates all unknown classes into a single category—GCD seeks to identify distinct unknown classes.

GCD methods typically facilitate knowledge transfer from labeled to unlabeled images using pseudo-labeling or semi-supervised learning approaches \cite{pu2023dccl,zhang2023promptcal,wen2023simgcd}. 
Other methods adopt semi-supervised learning objectives. The original GCD model \cite{vaze2022gcd} combines supervised contrastive loss with self-supervised loss, while PIM \cite{chiaroni2023pim_gcd} focuses on optimizing mutual information. CMS \cite{choi2024contrastive} employs mean-shift clustering alongside contrastive learning, and LegoGCD \cite{cao2024solving} incorporates Local Entropy Regularization and Dual-views KL divergence to enhance novel class differentiation and maintain performance on known classes.

\subsection{Object-Centric Contrastive Learning}

Contrastive learning seeks to enhance feature similarity among related entities while minimizing it among unrelated ones. While patch-level contrastive methods~\cite{VADeR, denseCL, PixPro} have been widely applied in semantic segmentation, they often overly focus on fine-grained details, neglecting broader object-level semantics.

To address this, object-level contrastive learning methods~\cite{maskcontrast, zadaianchuk2023unsupervised, cast, DetCon, hierarchicalGrouping, wen2022slotcon, seitzer2023bridging} aim to balance detailed perception with an object-centric perspective. For example, MaskContrast~\cite{maskcontrast} and COMUS~\cite{zadaianchuk2023unsupervised} leverage unsupervised saliency maps to guide pixel embeddings, while Odin~\cite{Odin} and DetCon~\cite{DetCon} utilize K-means clustering and heuristic masks, respectively. SlotCon~\cite{wen2022slotcon} introduces the concept of assigning pixels to “slots” for semantic representation learning, and DINOSAUR~\cite{seitzer2023bridging} further refines this by reconstructing pretrained feature embeddings, showing improved results across synthetic and real-world datasets.

\subsection{Adaptive Slot Attention}

Adaptive Slot Attention~\cite{fan2024adaptive} represents an advancement in object-centric learning by addressing the limitations of predefined slot numbers, a common constraint in existing methods like Slot Attention~\cite{locatello2020object}. Traditional models rely on fixed slot numbers, which can result in either over-segmentation or under-segmentation when the number of slots does not align with the complexity of the scene~\cite{engelcke2021genesis}.
GENESIS-V2~\cite{engelcke2021genesis} partially addresses this issue by employing a stochastic stick-breaking process to allow variable numbers of slots.

The Adaptive Slot Attention (AdaSlot) builds upon this foundation, introducing a novel complexity-aware framework to determine the number of slots dynamically. This is achieved using a lightweight slot selection module, leveraging the Gumbel-Softmax \cite{jang2022categorical} trick for differentiability and computational efficiency. Unlike prior methods, AdaSlot incorporates a masked slot decoder to suppress unselected slots during the decoding phase, thereby enhancing the quality of object representation and reconstruction.

\begin{figure*}[t]
    \centering
    \begin{minipage}{0.26\textwidth}
        \centering
        \includegraphics[width=\linewidth]{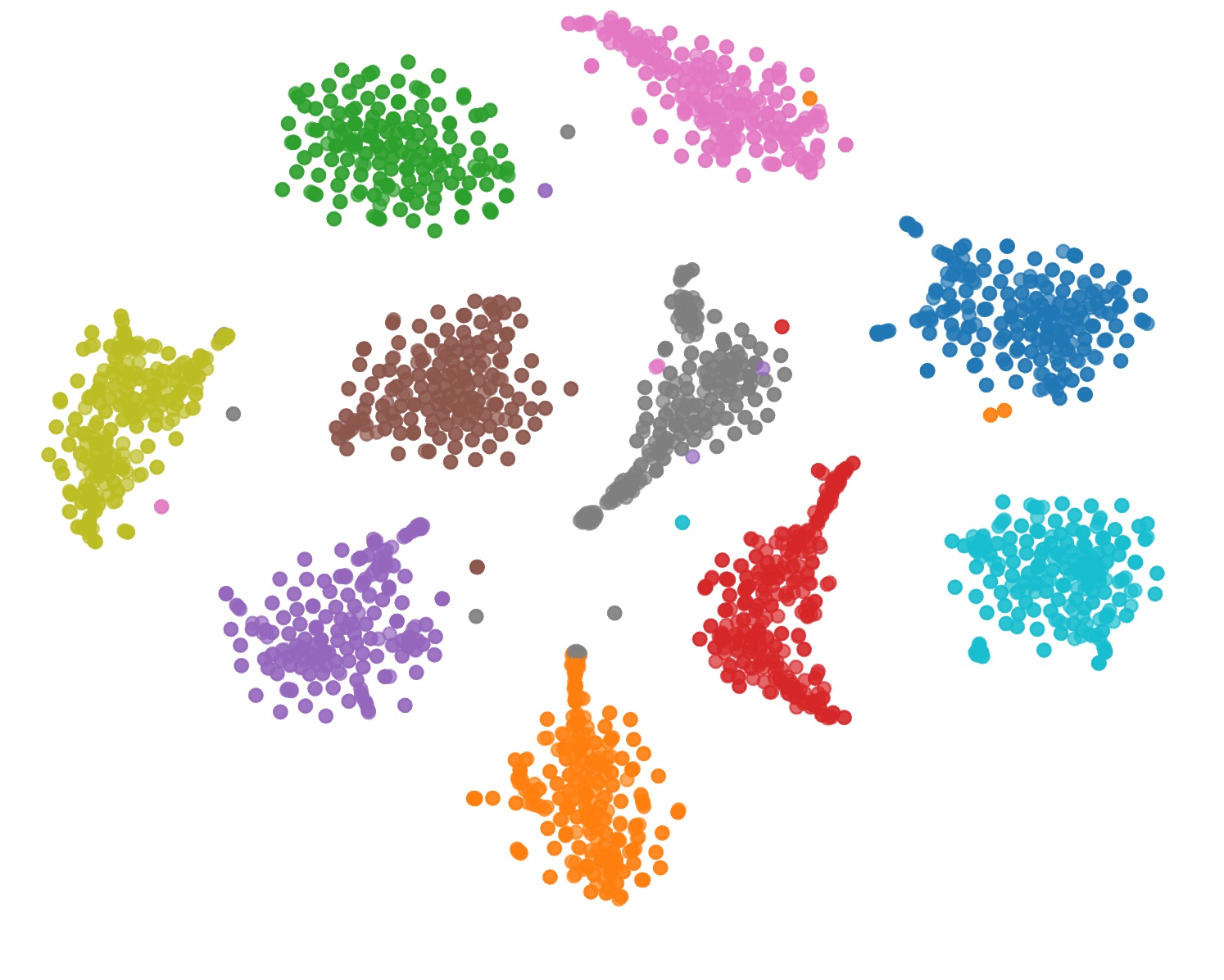}
        \textbf{(a)} CIFAR100
        \label{fig:tsne-CIFAR100}
    \end{minipage}
    \hfill
    \begin{minipage}{0.26\textwidth}
        \centering
        \includegraphics[width=\linewidth]{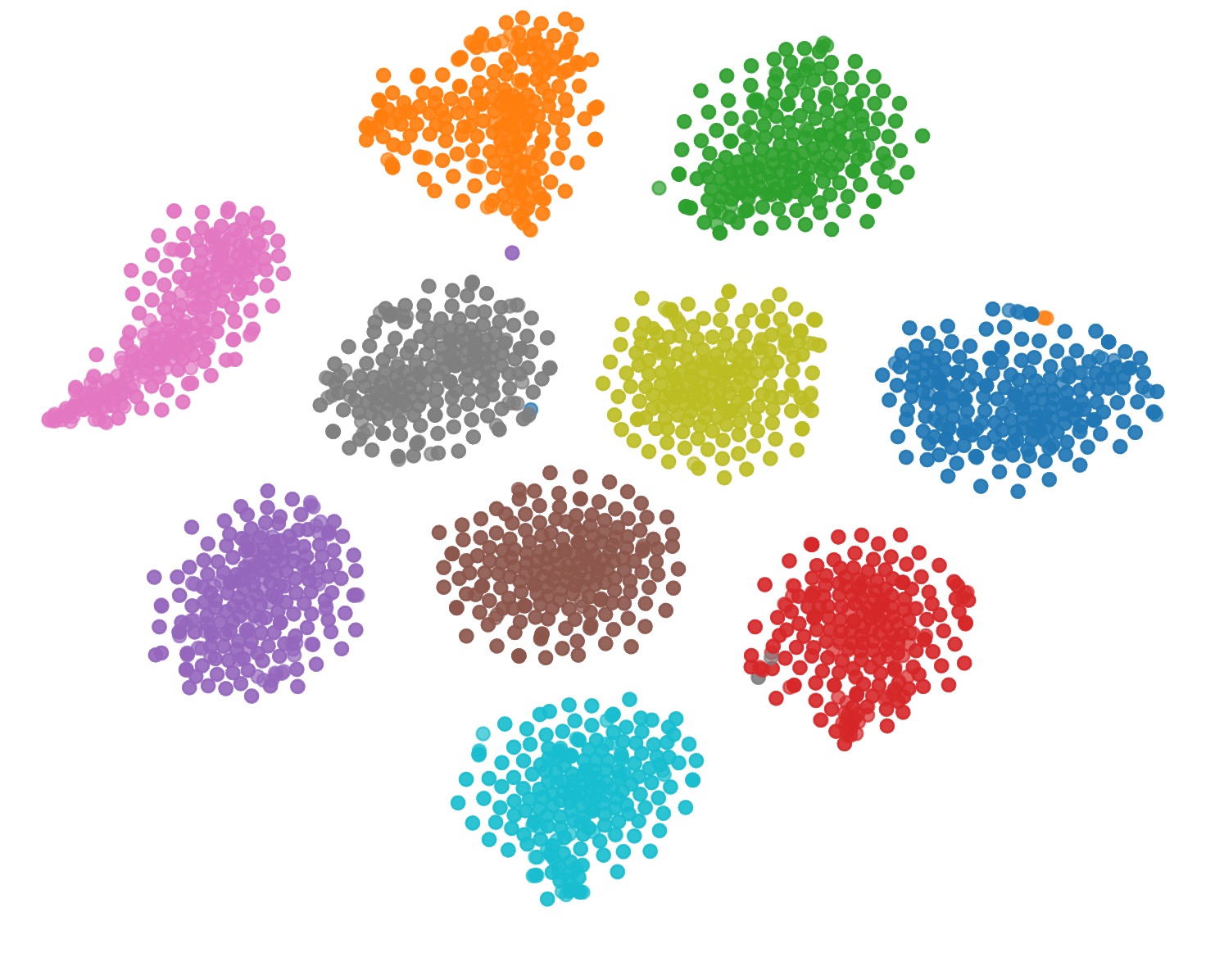}
        \textbf{(b)} ImageNet100
        \label{fig:tsne-imagenet100}
    \end{minipage}
    \hfill
    \begin{minipage}{0.26\textwidth}
        \centering
        \includegraphics[width=\linewidth]{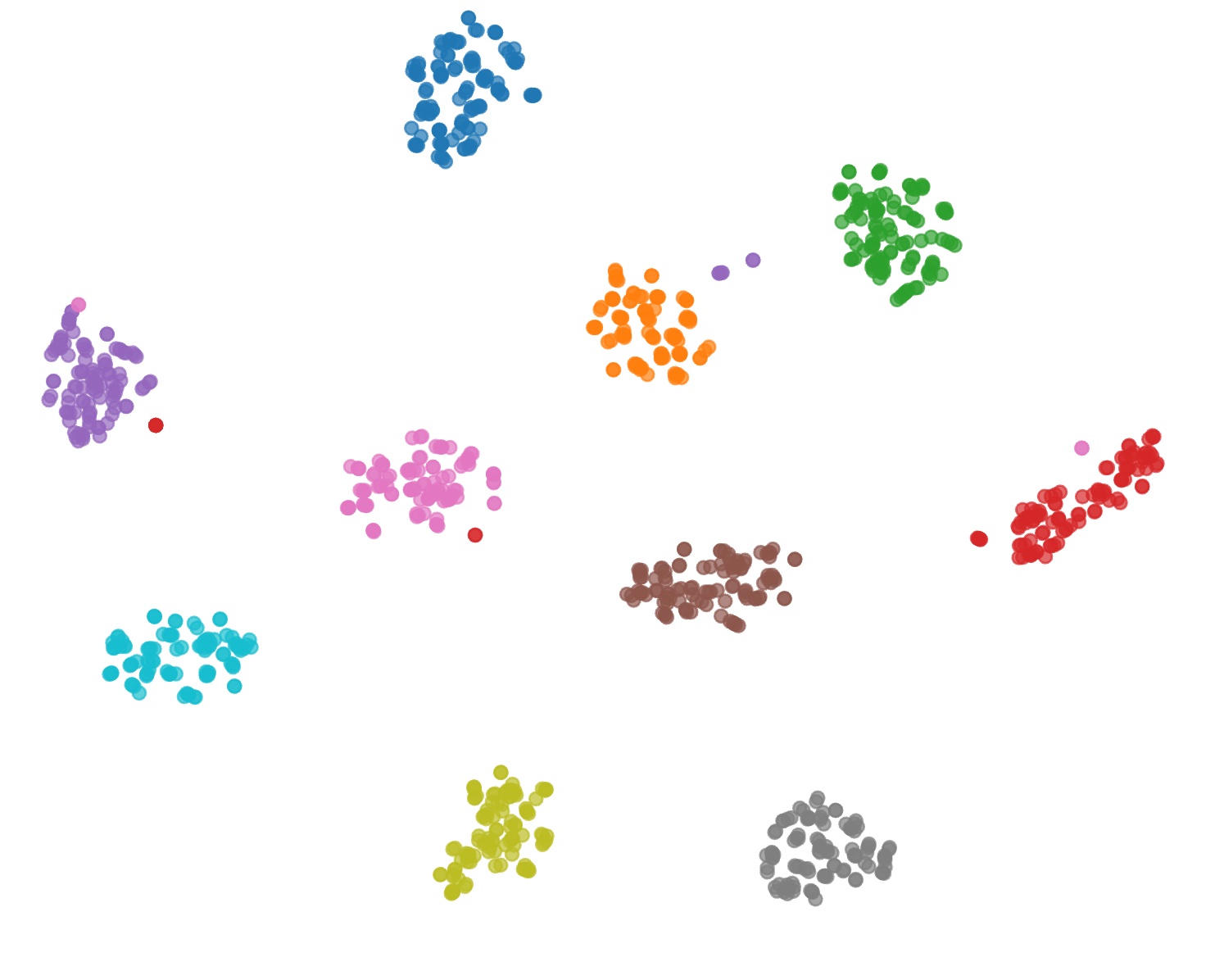}
        \textbf{(c)} CUB
        \label{fig:tsne-cub}
    \end{minipage}
    \vskip\baselineskip
    \begin{minipage}{0.26\textwidth}
        \centering
        \includegraphics[width=\linewidth]{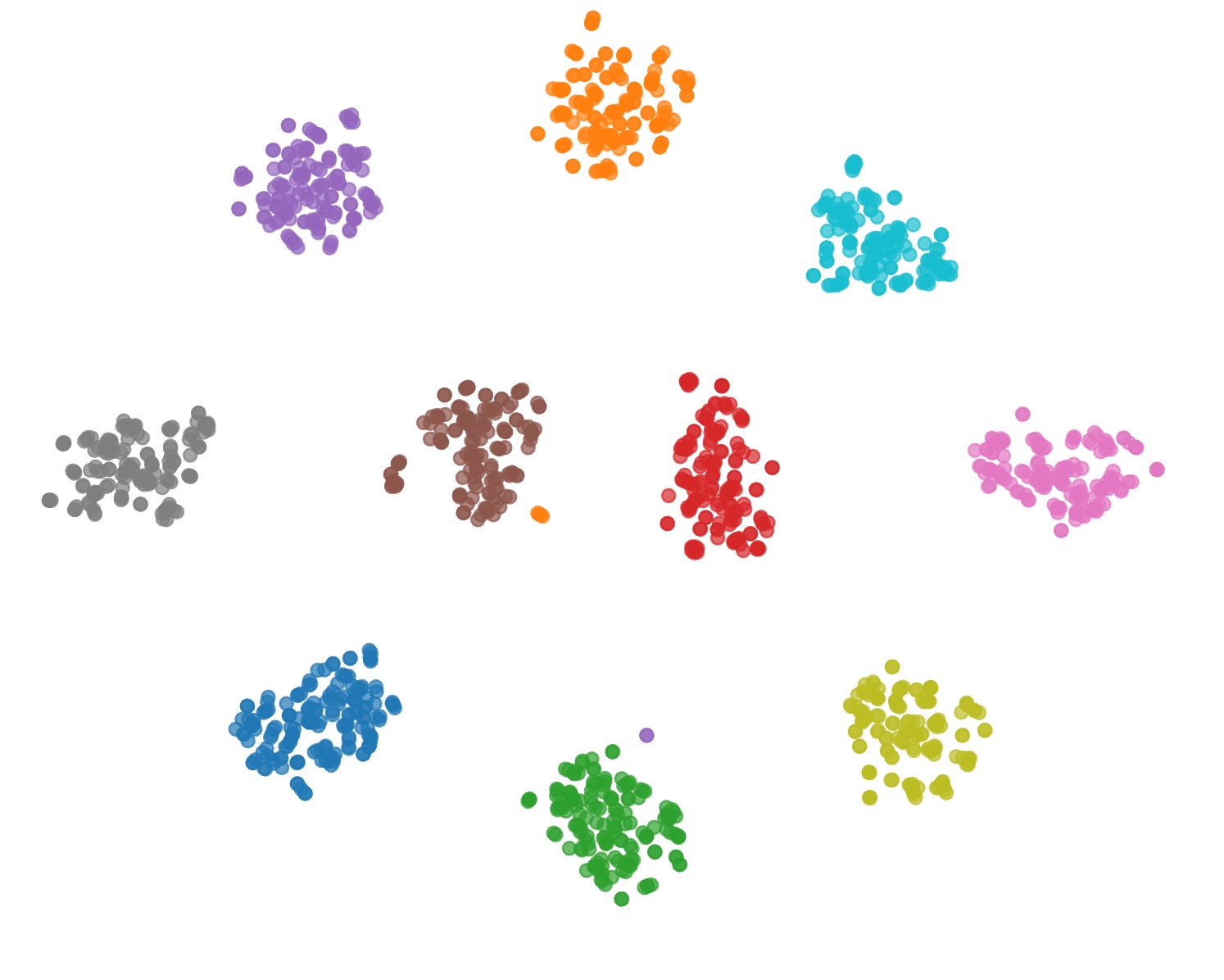}
        \textbf{(d)} Stanford Cars
        \label{fig:tsne-scar}
    \end{minipage}
    \hfill
    \begin{minipage}{0.26\textwidth}
        \centering
        \includegraphics[width=\linewidth]{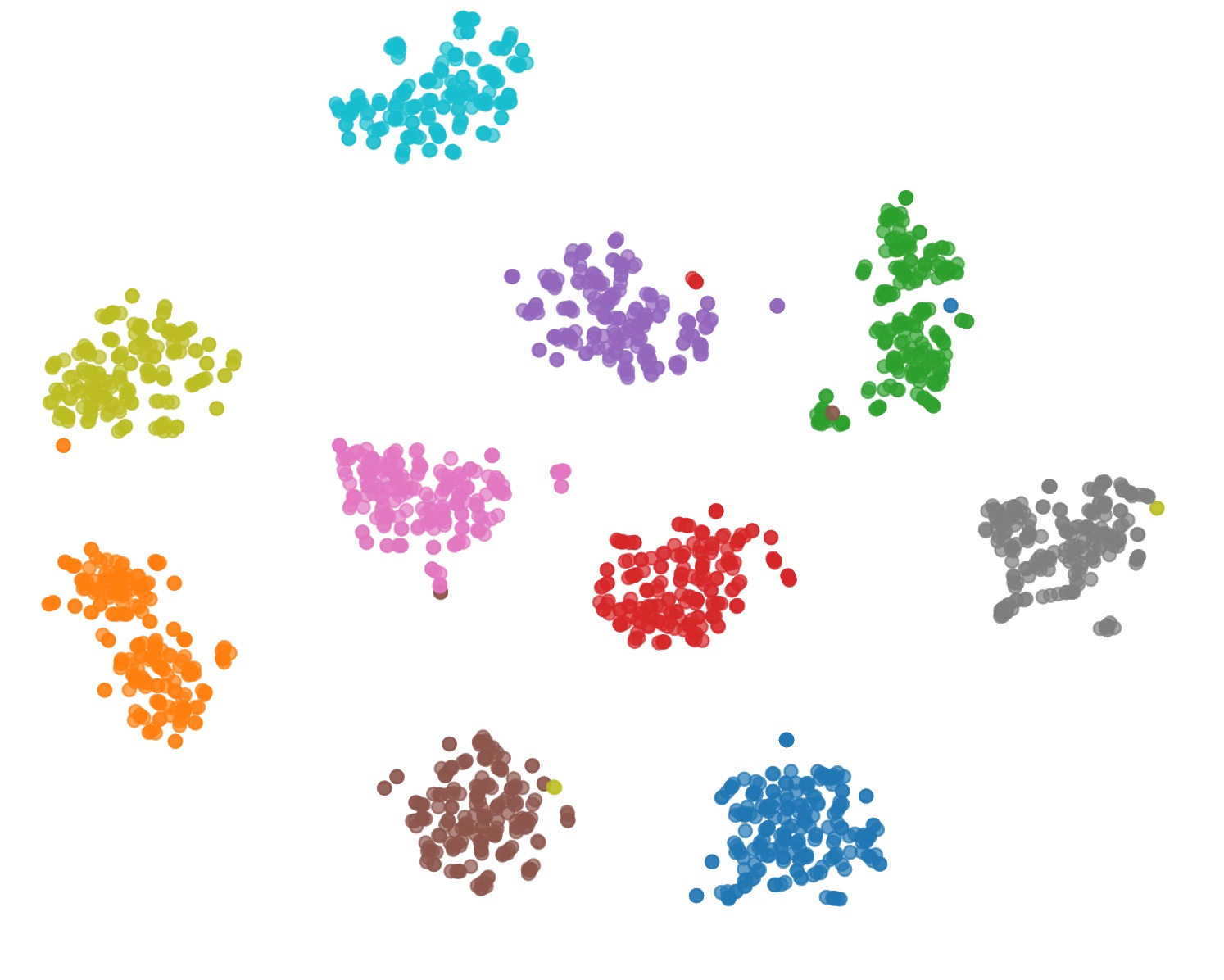}
        \textbf{(e)} FGVC Aircraft
        \label{fig:tsne-air}
    \end{minipage}
    \hfill
    \begin{minipage}{0.26\textwidth}
        \centering
        \includegraphics[width=\linewidth]{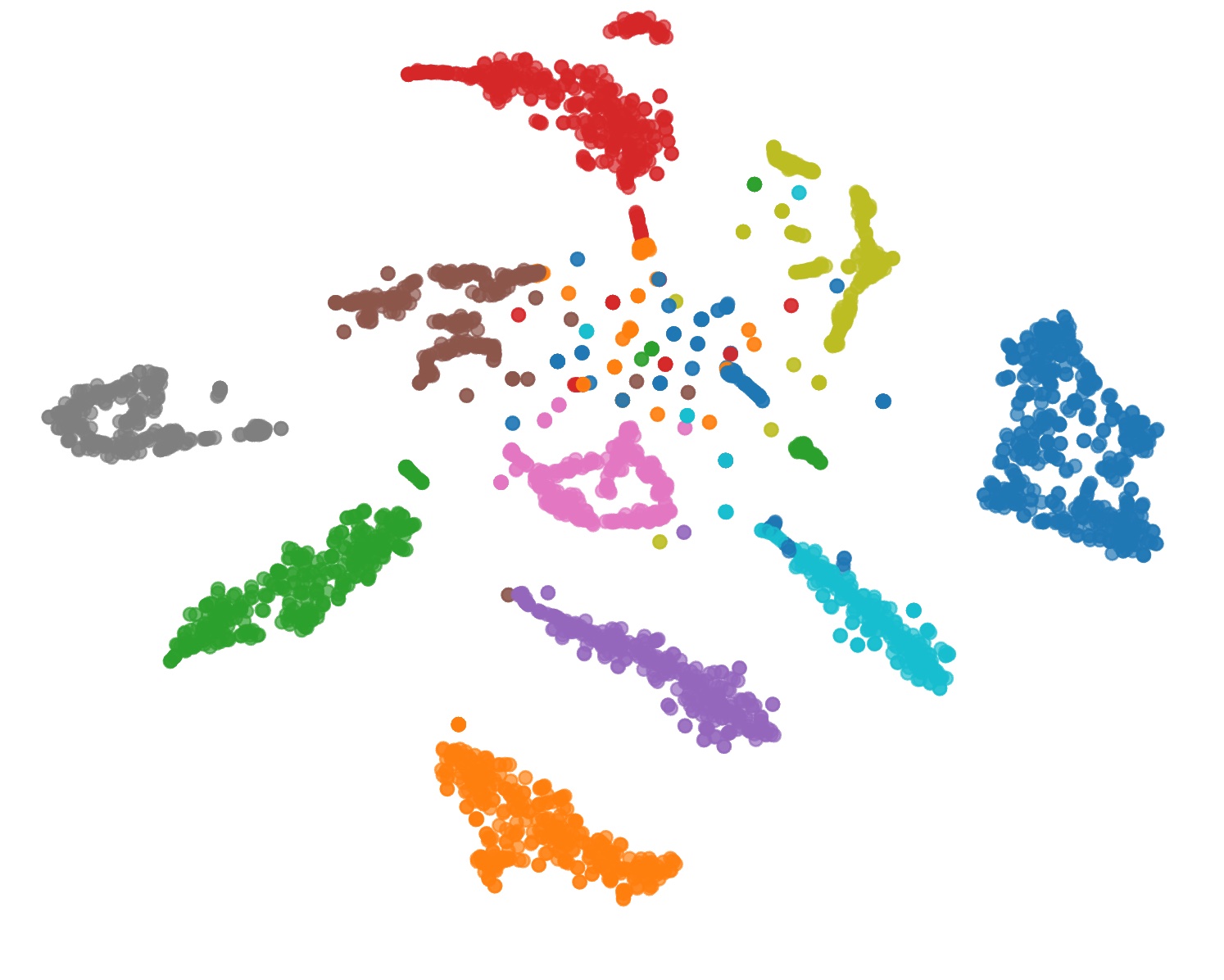}
        \textbf{(f)} Herbarium 19
        \label{fig:tsne-hr19}
    \end{minipage}
    \caption{Visualization of ten instances across six datasets using t-SNE \cite{van2008visualizing}. Each subfigure corresponds to a different dataset.}
    \label{fig:tsne-all}
    \vspace{-0.26cm}
\end{figure*}

\section{Method} \label{sec:method}

We leverage the spatial local features of objects for GCD by introducing cluster-centric contrastive learning. To incorporate these spatial local features, we extract local embeddings from the DINO encoder and analyze their relationships using the proposed Adaslot. Slots are designed to capture clustered features, such as semantic categories, contours, and shapes, and are configured to be sample-specific. These slots are supervised using a reconstruction loss (Sec.~\ref{subsec:lowlevel}). 
Subsequently, image-scale features are derived from the component clusterer, and the framework is trained through image-scale contrastive learning (Sec.~\ref{subsec:overall}, Fig.~\ref{fig:framework}). Further preliminaries are provided in the supplemental material.

\subsection{Clustered Features with Component Clusterer}
\label{subsec:lowlevel}

Since spatial local information plays a pivotal role in distinguishing between categories, we focus on extracting inherent clustered features for GCD. These features are difficult to represent through global information alone, so we propose learning a component clusterer $\boldsymbol{S}$ to group pixels into meaningful clusters. Within each cluster, pixels share semantic consistency (e.g., similar clustered feature representations such as texture or color), while pixels from different clusters exhibit semantic incoherence.

As illustrated in Fig.~\ref{fig:framework}, clustered features are obtained through a DINO encoder $f_{\theta}$ and a component clusterer module $\mathcal{S}_{\theta}$. Given an input image $\boldsymbol{x}$, the encoder $f_{\theta}$ processes it into a spatial local feature map $\boldsymbol{h} \in \mathbb{R}^{H \times W \times D}$. This feature map is then passed to the component clusterer module $\mathcal{S}_{\theta}$, yielding the clustered features $\boldsymbol{s}_{out} = \mathcal{S}_{\theta}\left(\boldsymbol{s}_{init}, \boldsymbol{h}, \boldsymbol{h}\right) \in \mathbb{R}^{K_{max} \times D}$, where $K_{max}$ represents the maximum number of slots in the component clusterer, and $\boldsymbol{s}_{init}$ denotes the initial slots. By initializing these slots with random noise, the module gains greater flexibility in capturing sample-specific clustered features. 

Our component clusterer module $\mathcal{S}_{\theta}$ builds upon the Adaslot~\cite{fan2024adaptive}, with the initial slots $\boldsymbol{s}_{init} \in \mathbb{R}^{K_{max} \times D}$ initialized from a Gaussian distribution to ensure sample specificity. Inspired by previous work in low-level domains, we hypothesize that reconstruction loss is effective for learning clustered features. To this end, we employ a masked slot decoder~\cite{fan2024adaptive}, denoted as $d_{\theta}$, to reconstruct $\boldsymbol{s}_{out}$ back to the original feature map $\boldsymbol{\hat{h}}$. We use a simple Mean Squared Error (MSE) loss to train this low-level branch:
\begin{equation} \label{eq:mse}
\mathcal{L}_{\theta}^{\mathrm{rec}} = ||\boldsymbol{h} - \boldsymbol{\hat{h}}||
\end{equation}

\subsection{Cluster-Centric Contrastive Learning}
\label{subsec:overall}

To derive an image-scale representation from the slot-scale features $\boldsymbol{s}^{i}_{out}$, we employ a pooling method that computes the average of $\boldsymbol{s}^{i}_{k}$. This operation aggregates the slot-scale features into a single image-scale representation.

For the component clusterers, representations $\boldsymbol{g}$ can be extracted independently. These pooled features are then concatenated with the original global feature $\boldsymbol{g}_{dino} \in \mathbb{R}^{D}$ obtained from the DINO encoder. This concatenation results in a unified vector $\boldsymbol{g}_{all} \in \mathbb{R}^{3D}$, which incorporates both global semantic information and clustered features. 

The unified vector serves as the image-scale representation and is passed to the classification head for both supervised image-scale contrastive learning, $\mathcal{L}_{\theta}^{\mathrm{IC(S)}}$, and unsupervised contrastive learning, $\mathcal{L}_{\theta}^{\mathrm{IC(U)}}$, following the CMS implementation~\cite{choi2024contrastive}.

The overall optimization involves jointly minimizing the low-level reconstruction loss (Eq.~\ref{eq:mse}) and the image-scale contrastive learning losses. These are balanced using weighting factors $\lambda$ as follows:

\begin{equation}
\begin{aligned}
    \mathcal{L}_{\theta}^{\mathrm{Overall}} &= \lambda_{rec}\mathcal{L}_{\theta}^{\mathrm{rec}} + \mathcal{L}_{\theta}^{\mathrm{IC}}, \\
\mathcal{L}_{\theta}^{\mathrm{IC}} &= \lambda_{IC}^{S}\mathcal{L}_{\theta}^{\mathrm{IC(S)}} + \lambda_{IC}^{U}\mathcal{L}_{\theta}^{\mathrm{IC(U)}}.
\end{aligned}
\end{equation}

\section{Experiments}

\begin{table*}[t!]
    \centering
        \caption{Comparison with the state-of-the-art methods on GCD.}
    \tabcolsep=0.1cm  %
    \resizebox{2.0\columnwidth}{!}{
    \begin{tabular}[b]{>{\raggedright\arraybackslash}p{0.21\textwidth} ccc ccc ccc ccc ccc ccc}
    \toprule
        \multirow{2.5}{*}{Method} &
        \multicolumn{3}{c}{CIFAR100 \cite{Krizhevsky09cifar}} &
        \multicolumn{3}{c}{ImageNet100 \cite{deng09imagnet}} &
        \multicolumn{3}{c}{CUB \cite{cub200}} &
        \multicolumn{3}{c}{Stanford Cars \cite{Cars196}} &
        \multicolumn{3}{c}{FGVC Aircraft \cite{maji2013aircraft}} &
        \multicolumn{3}{c}{Herbarium 19 \cite{Chuan19herbarium}} \\
        \cmidrule(lr){2-4} \cmidrule(lr){5-7} \cmidrule(lr){8-10}
        \cmidrule(lr){11-13} \cmidrule(lr){14-16} \cmidrule(lr){17-19}
        & All & Old & New & All & Old & New & All & Old & New
        & All & Old & New & All & Old & New & All & Old & New\\
    \midrule
        RankStats+~\cite{Han2020rs+} (ICLR'20) & 58.2 & 77.6 & 19.3 & 37.1 & 61.6 & 24.8 & 33.3 & 51.6 & 24.2 & 28.3 & 61.8 & 12.1 & 26.9 & 36.4 & 22.2 & 27.9 & 55.8 & 12.8 \\
        UNO+~\cite{fini2021uno} (ICCV'21) & 69.5 & 80.6 & 47.2 & 70.3 & 95.0 & 57.9 & 35.1 & 49.0 & 28.1 & 35.5 & 70.5 & 18.6 & 40.3 & 56.4 & 32.2 & 28.3 & 53.7 & 14.7 \\
        ORCA~\cite{cao2022orca} (CVPR'22) & 69.0 & 77.4 & 52.0 & 73.5 & 92.6 & 63.9 & 35.3 & 45.6 & 30.2 & 23.5 & 50.1 & 10.7 & 22.0 & 31.8 & 17.1 & 20.9 & 30.9 & 15.5\\
        GCD~\cite{vaze2022gcd} (CVPR'22) & 73.0 & 76.2 & 66.5 & 74.1 & 89.8 & 66.3 & 51.3 & 56.6 & 48.7 & 39.0 & 57.6 & 29.9 & 45.0 & 41.1 & 46.9 & 35.4 & 51.0 & 27.0\\
        DCCL~\cite{pu2023dccl} (CVPR'23) & 75.3 & 76.8 & 70.2 & 80.5 & 90.5 & 76.2 & 63.5 & 60.8 & \underline{64.9} & 43.1 & 55.7 & 36.2 & - & - & - & - & - & - \\
        PromptCAL~\cite{zhang2023promptcal} (CVPR'23) & 81.2 & 84.2 & 75.3 & 83.1 & 92.7 & 78.3 & 62.9 & 64.4 & 62.1 & 50.2 & 70.1 & 40.6 & 52.2 & 52.2 & \textbf{52.3} & 37.0 & 52.0 & 28.9\\
        GPC~\cite{zhao2023gmm_gcd} (ICCV'23) & 77.9 & 85.0 & 63.0 & 76.9 & 94.3 & 71.0 & 55.4 & 58.2 & 53.1 & 42.8 & 59.2 & 32.8 & 46.3 & 42.5 & 47.9 & - & - & - \\
        SimGCD~\cite{wen2023simgcd} (ICCV'23) & 80.1 & 81.2 & 77.8 & 83.0 & 93.1 & 77.9 & 60.3 & 65.6 & 57.7 & 53.8 & 71.9 & 45.0 & 54.2 & 59.1 & 51.8 & 44.0 & \underline{58.0} & 36.4 \\
        PIM~\cite{chiaroni2023pim_gcd} (ICCV'23) & 78.3 & 84.2 & 66.5 & 83.1 & \underline{95.3} & 77.0 & 62.7 & \underline{75.7} & 56.2 & 43.1 & 66.9 & 31.6 & - & - & - & 42.3 & 56.1 & 34.8\\
        CMS~\cite{choi2024contrastive} (CVPR'24) & \underline{82.3} & \textbf{85.7} & 75.5 & 84.7 & \textbf{95.6} & 79.2 & \underline{68.2} & \textbf{76.5} & 64.0 & 56.9 & \textbf{76.1} & \underline{47.6} & \underline{56.0} & \textbf{63.4} & \textbf{52.3} & 36.4 & 54.9 & 26.4 \\
        LegoGCD~\cite{cao2024solving} (CVPR'24) & 81.8 & 81.4 & \textbf{82.5} & \underline{86.3} & 94.5 & \textbf{82.1} & 63.8 & 71.9 & 59.8 & \underline{57.3} & \underline{75.7} & \textbf{48.4} & 55.0 & 61.5 & 51.7 & \underline{45.1} & 57.4 & \underline{38.4}\\
        CDAD-NET~\cite{rongali2024cdad} (CVPRW'24) & 75.6 & 76.3 & \underline{80.0} & 80.5 & 80.8 & \underline{80.3} & - & - & - & - & - & - & - & - & - & - & - & -\\
        \ccol \shortmethod \ (Ours) & \ccol \textbf{83.4} & \ccol \underline{85.3} & \ccol 76.2 & \ccol \textbf{87.0} & \ccol 95.2 & \ccol 78.7 & \ccol \textbf{71.4} & \ccol 75.2 & \ccol \textbf{67.6} & \ccol \textbf{59.6} & \ccol \underline{75.7} & \ccol 44.2 & \ccol \textbf{56.4} & \ccol \textbf{63.4} & \ccol 49.4 & \ccol \textbf{50.1} & \ccol \textbf{59.6} & \ccol \textbf{39.9} \\

    \midrule  

    \bottomrule
    \end{tabular}%
    }

    \label{table:gcd}
    \vspace{-0.3cm}
\end{table*}%

\subsection{Experimental Setup}
\begin{table}[b]
\centering
\vspace{-0.1cm}
\caption{Overview of datasets used in our experiments. Each dataset is described by the number of known and unknown classes, as well as the sizes of the labeled and unlabeled subsets.}
\begin{tabular}{lcccccc}
\toprule
\multirow{2}{*}{Dataset} & \multicolumn{2}{c}{Classes} & \multicolumn{2}{c}{Images} \\
\cmidrule(lr){2-3}\cmidrule(lr){4-5}
& labeled & unlabeled & labeled & unlabeled \\
\midrule
CIFAR100 \cite{Krizhevsky09cifar}   & 80   & 100   & 20k    & 30k \\
ImageNet100 \cite{deng09imagnet} & 50   & 100   & 31.9k  & 95.3k \\
CUB \cite{cub200}       & 100  & 200   & 1.5k   & 4.5k \\
SCars \cite{Cars196}     & 98   & 196   & 2.0k   & 6.1k \\
FGVC \cite{maji2013aircraft}      & 50   & 100   & 1.7k   & 5.0k \\
Herb19 \cite{Chuan19herbarium}    & 341  & 683   & 8.9k   & 25.4k \\
\bottomrule
\end{tabular}
\label{tab:datasets}
\end{table}

\smallbreakparagraph{Datasets.}
We conduct an evaluation of the proposed method on six distinct datasets. For each dataset, we begin by selecting a subset of classes with known labels from the training set. Half of the images from these labeled classes, amounting to 50\%, are sampled to form the labeled dataset, $\mathcal{D_L}$. The remaining samples from these classes, along with all samples from the unlabeled classes, are used to create the unlabeled dataset, $\mathcal{D_U}$. Additionally, a validation dataset is derived for the labeled classes using the corresponding test or validation split from each dataset.
Table~\ref{tab:datasets} summarizes the dataset splits used in our experiments.
More details about the datasets are provided in the supplemental material.

\subsection{Training Details}

For the image encoder, we utilize the pre-trained DINO ViT-B/16 model \cite{dino}. To ensure fair comparisons with other methods \cite{vaze2022gcd, zhang2023promptcal, pu2023dccl}, we include a projection head in the architecture. In this configuration, only DINO’s final layer is trainable, while the remaining layers are frozen. The component clusterer module and projection head, however, are fully trainable.
All experiments are conducted on a single RTX-A6000 GPU. 
Following \cite{vaze2022gcd}, we assign loss weights of $0.6$, $0.3$, and $0.1$ to the unsupervised learning loss, supervised learning loss, and MSE loss, respectively.

\smallbreakparagraph{Component Clusterer Module.}
The component clusterer module defines the clustered feature dimensionality, $D_{clustered}$, as $64$, with a maximum of $50$ slots. The clustered features decoder comprises a four-layer MLP with ReLU activations, designed to reconstruct DINO’s local features. The output dimensionality is $D_{feat} + 1$, with the additional dimension allocated for the alpha value. Each hidden layer in the MLP has a dimensionality of $128$. A learnable positional encoding of size $N \times D_{clustered}$ is applied and broadcasted across the slots to match the number of patches, $N$.

\begin{table*}[t!]
        \centering
        \caption{Performance with different loss ratios across datasets.}
        \tabcolsep=0.20cm
        \scalebox{0.90}{%
        \begin{tabular}[t]{l  c  c c | ccc ccc ccc ccc ccc ccc}
        \toprule
         &
         &
         &
         &
        \multicolumn{3}{c}{CIFAR100}&
        \multicolumn{3}{c}{ImageNet100} &
        \multicolumn{3}{c}{CUB}&
        \multicolumn{3}{c}{Stanford Cars}&
        \multicolumn{3}{c}{FGVC Aircraft }&
        \multicolumn{3}{c}{Herbarium 19}
        \\
        \cmidrule(lr){2-2} \cmidrule(lr){3-3}
        \cmidrule(lr){4-4}  \cmidrule(lr){5-7} \cmidrule(lr){8-10} \cmidrule(lr){11-13} \cmidrule(lr){14-16} \cmidrule(lr){17-19} \cmidrule(lr){20-22} 
         & $\lambda_{IC}^{U}$ & $\lambda_{IC}^{S}$ & $\lambda_{rec}$ & All & Old & New & All & Old & New & All & Old & New & All & Old & New& All & Old & New& All & Old & New\\
        \midrule
        \ccol(1) & \ccol 0.6 & \ccol 0.3 & \ccol 0.1 & \ccol 83.4 & \ccol 85.3 & \ccol 76.2 & \ccol \textbf{87.0} & \ccol 95.2 & \ccol \textbf{78.7} & \ccol \textbf{71.4} & \ccol 75.2 & \ccol \textbf{67.6} & \ccol \textbf{59.6} & \ccol \textbf{75.7} & \ccol \textbf{44.2} & \ccol \textbf{56.4} & \ccol \textbf{63.4} & \ccol \textbf{49.4} & \ccol \textbf{50.1} & \ccol \textbf{59.6} & \ccol \textbf{39.9} \\
        (2) & 0.55 & 0.25 & 0.2 &83.1 &\textbf{85.6} &73.0 & 86.5 &95.1 & 77.9 & 68.5 & 75.9 &61.2 & 47.5& 68.0&27.8& 53.8& 62.0&45.6&47.5 &55.4 &39.2\\
        (3) & 0.5 & 0.2 & 0.3  & 83.4& 85.3&75.6&86.6 &95.1 & 78.0& 69.2 & \textbf{76.4}& 62.0& 48.1&68.4 &28.4&52.8 & 61.8&43.8& 46.9& 55.3&37.9\\
        (4) & 0.45 & 0.15 & 0.4  & \textbf{83.5} & 85.3 &\textbf{76.6} &86.8 &\textbf{95.4} &78.1 &69.1 &75.1 & 63.1&46.2 &66.4 &26.7&51.3 &57.4 &45.2&46.3 &53.7 &38.5\\
        (5) & 0.4 & 0.1 & 0.5  & 83.1 & 85.0&75.6 &86.8 &95.2 &78.4 &67.4 &73.9 & 61.0&41.7 &61.5 &22.6&47.5 &57.4 &37.7&45.2 &52.4 &37.6\\
        \bottomrule
        \end{tabular}%
        }

        \label{tab:ablation_loss_weight}
        \vspace{-0.3cm}
\end{table*}

\begin{table}[b]
        \centering
                \caption{Experiments under different choices of $K_{max}$ on the CUB dataset.}
                \label{tab:ablation_k_max}
      \vspace{-0.1cm}
        \begin{tabular}[t]{l  c | ccc}
        \toprule
             & $K_{max}$ & All & Old & New\\
        \midrule
        \ccol (1) & \ccol 50 & \ccol \textbf{71.4} & \ccol 75.2 & \ccol \textbf{67.6} \\
        (2) & 40 & 70.9 & \textbf{76.9} & 64.9 \\
        (3) & 30 & 70.2 & 76.2 & 64.3 \\
        (4) & 20 & 70.0 & 76.3 & 63.7 \\
        (5) & 10 & 69.2 & 75.5 & 62.9 \\
        (6) & 5 & 70.3 & 75.4 & 65.3 \\
        \midrule
        (7) & 5 (fixed) & 69.7 & 76.5 & 62.9 \\
        \bottomrule
        \end{tabular}%

        \label{tab:ablation_slot}

\end{table}

\smallbreakparagraph{Evaluation.}
Performance is evaluated by clustering all images in the dataset $\mathcal{D}$ and measuring accuracy on the unlabeled set $\mathcal{D}_{\text{UL}}$. Consistent with \cite{vaze2022gcd}, accuracy is calculated using the Hungarian matching method~\cite{kuhn1955hungarian}, which aligns cluster assignments with ground-truth labels based on instance overlap between class pairs. Classes without a match are marked as incorrect, while the most frequent class in each ground-truth cluster is considered correct. We report accuracy across all unlabeled data (“All”) and separately for known classes (“Old”) and unknown classes (“New”), as shown in Table \ref{table:gcd}.

More training details are provided in the supplemental material.

\subsection{Main Results}

Our method achieves consistently strong performance across a variety of GCD benchmarks, often surpassing or matching state-of-the-art approaches. By effectively integrating spatial local features with global context, the model demonstrates adaptability across diverse categories, from general objects to fine-grained and specialized domains. A t-SNE visualization, as depicted in Fig.~\ref{fig:tsne-all}, shows how ten samples from each of the six datasets form distinct clusters based on their learned representations.

\smallbreakparagraph{Performance on Herbarium 19.}  
On the challenging Herbarium 19 dataset, our method outperforms a leading competitor by more than four percentage points. This significant improvement underscores the critical role of leveraging local details to distinguish subtle characteristics in plant species. The model’s component clusterer module excels in capturing these fine-grained variations while maintaining a holistic understanding of each sample.

\smallbreakparagraph{Performance on Fine-Grained Classification.}  
For fine-grained tasks such as CUB, Stanford Cars, and FGVC Aircraft, our approach exhibits clear advantages by capturing intricate details through component clusterers. While some specialized architectures may achieve higher scores on specific benchmarks, our method remains highly competitive overall. This indicates that the synergy of spatial local and global features is particularly effective for recognizing subtle differences in visually similar categories.

\smallbreakparagraph{Performance on Object Recognition Benchmarks.}  
On general object recognition benchmarks like CIFAR100 and ImageNet100, our model consistently delivers robust results, outperforming strong baselines by a clear margin. This demonstrates the framework’s capacity to handle broad visual diversity, validating the effectiveness of cluster-centric contrastive learning in adapting to both generic and domain-specific tasks.

\subsection{Ablation Studies}

\smallbreakparagraph{Influence of $\lambda$.}  
Our ablation study examines how varying the loss weights \(\lambda_{IC}^U\), \(\lambda_{IC}^S\), and \(\lambda_{rec}\) affects the trade-off between retaining old class knowledge and adapting to new classes, as shown in Table \ref{tab:ablation_loss_weight}. Lowering \(\lambda_{IC}^U\) and \(\lambda_{IC}^S\) shifts the model's focus to reconstruction (\(\lambda_{rec}\)), boosting representational consistency but slightly reducing new class performance. Conversely, increasing these classification-oriented weights improves adaptation to new classes at the cost of minor degradation in old class performance. Proper tuning of these hyperparameters enables a balanced trade-off between learning new data and preserving previously learned accuracy.

\smallbreakparagraph{Influence of $K_{max}$.}  
We also examine the impact of varying the maximum slot capacity \(\,K_{max}\) on the CUB dataset, as presented in Table \ref{tab:ablation_k_max}. The study evaluates multiple configurations of \(\,K_{max}\), from larger capacities to more restrictive ones. Across these settings, the model demonstrates stable performance on both old and new categories, highlighting its robustness to variations in \(\,K_{max}\). Interestingly, even with lower slot capacities, the model outperforms a comparable fixed-slot baseline, indicating that flexibility in the number of slots enhances incremental learning while minimally affecting previously learned knowledge.

\section{Conclusion}
In this paper, we introduced a cluster-centric contrastive learning framework for Generalized Category Discovery (GCD) that effectively leverages both global and spatial local features. By integrating Adaptive Slot Attention (AdaSlot), our method dynamically determines the number of slots needed for semantic decomposition of the spatial local feature maps, thus eliminating the rigid requirement of predefined slot counts. This integration enables more expressive and fine-grained representations, improving open-world class discovery. Extensive experiments on public and fine-grained datasets illustrate the efficacy of our approach, demonstrating how spatial local information can be harnessed to enhance category discovery.

\bibliographystyle{ieeetr}
\bibliography{main}

\end{document}